\documentclass[runningheads]{llncs}
\usepackage[T1]{fontenc}
\usepackage{graphicx}
\usepackage{marvosym}
\usepackage{times}
\usepackage{epsfig}
\usepackage{setspace}
\usepackage{array}
\usepackage{multirow}
\usepackage{booktabs}
\usepackage{color}
\usepackage{amsfonts}
\usepackage{lineno}
\usepackage{threeparttable}
\usepackage{amsmath,bm}
\usepackage{algorithm}
\usepackage{algpseudocode}
\usepackage{siunitx}
\usepackage{lineno,hyperref}
\usepackage{pfnote}

\begin{document}
%
%
\titlerunning{Location embedding based pairwise distance learning}
%

\title{Location embedding based pairwise distance learning for fine-grained diagnosis of urinary stones}
\author{Qiangguo Jin\inst{1,2} \and
Jiapeng Huang\inst{1} \and
Changming Sun\inst{3} \and
Hui Cui\inst{4} \and
Ping Xuan \inst{5} \and
Ran Su\inst{6} \and 
Leyi Wei\inst{7,8} \and 
Yu-Jie Wu\inst{9,10} \and
Chia-An Wu\inst{9} \and
Henry B.L. Duh\inst{11} \and
Yueh-Hsun Lu\textsuperscript{\Letter}\inst{9,12}}
\authorrunning{Qiangguo Jin et al.}
\institute{School of Software, Northwestern Polytechnical University, Shaanxi, China \and 
Yangtze River Delta Research Institute of NPU, Suzhou, China \and
CSIRO Data61, Sydney, Australia \and
Department of Computer Science and Information Technology, La Trobe University, Melbourne, Australia \and
School of Engineering, Shantou University, Guangdong, China \and
College of Intelligence and Computing, Tianjin University, Tianjin, China \and
School of Information, Xiamen University, Xiamen, China \and
AIDD, Faculty of Applied Science, Macao Polytechnic University, Macao SAR, China \and
Department of Radiology, Shuang-Ho Hospital, Taipei Medical University, Taipei, Taiwan \and
Master of Public Health Program, National Yang Ming Chiao Tung University, Taipei, Taiwan \and
School of Design, The Hong Kong Polytechnic University, Hong Kong \and
Department of Radiology, School of Medicine, College of Medicine, Taipei Medical University, Taipei, Taiwan\\\email{20001@s.tmu.edu.tw}
}

%
%
%
\maketitle              

\begin{abstract}
The precise diagnosis of urinary stones is crucial for devising effective treatment strategies. The diagnostic process, however, is often complicated by the low contrast between stones and surrounding tissues, as well as the variability in stone locations across different patients. To address this issue, we propose a novel location embedding based pairwise distance learning network (LEPD-Net) that leverages low-dose abdominal X-ray imaging combined with location information for the fine-grained diagnosis of urinary stones. LEPD-Net enhances the representation of stone-related features through context-aware region enhancement, incorporates critical location knowledge via stone location embedding, and achieves recognition of fine-grained objects with our innovative fine-grained pairwise distance learning. Additionally, we have established an in-house dataset on urinary tract stones to demonstrate the effectiveness of our proposed approach. Comprehensive experiments conducted on this dataset reveal that our framework significantly surpasses existing state-of-the-art methods.

\end{abstract}
\section{Introduction}
Urinary tract stones (UTS) are a primary cause of low back pain, presenting a significant diagnostic challenge in healthcare. Although rarely fatal, their prevalence is remarkably high, affecting up to 10\% of the population in developed countries~\cite{khan2016kidney,scales2012prevalence}. Non-contrast computed tomography (NCCT) is established as the gold standard for urolithiasis diagnosis, demonstrating a diagnostic accuracy exceeding 92\%~\cite{kobayashi2021computer}. However, due to the associated costs and radiation exposure from NCCT, low-dose abdominal X-rays, or KUB (Kidney, Ureter, and Bladder) radiography, are considered a viable, cost-effective initial diagnostic option for UTS diagnosis. Yet, the diagnostic accuracy of KUB exams falls significantly short of NCCT, with reported accuracies ranging from 44\% to 77\%~\cite{turk2016eau}. Therefore, the development of automated and precise methods for diagnosing UTS from cost-effective KUB images is worth investigating.

However, diagnosing UTS from KUB images presents several distinct challenges. Firstly, distinguishing small-scale UTS from high-density objects like large intestinal feces, vascular calcifications, and phleboliths within KUB images proves to be intricate. Secondly, phleboliths, or small vein wall calcifications commonly found in the pelvis, occurring in approximately 40\% of adults~\cite{luk2017pelvic}, are often located near the ureters, making it especially challenging to distinguish them from ureter stones. Hence, the significant variability in stone locations across patients greatly complicates the diagnostic task. Finally, the uneven distribution and limited sample sizes of various stone types (as shown in Fig.~\ref{data_introduction}(a)), alongside the low contrast between them, pose significant obstacles to conducting a fine-grained diagnosis. Thus, the development of more specialized methods is needed to address these diagnostic challenges.

Deep learning has attracted substantial research interest across various fields of medical diagnosis~\cite{han2022radiomics,zhou2023self,zhou2023novel,lu2023hacl}. For example, Zhang~et~al.~\cite{zhang2019attention} introduced an attention residual learning convolutional neural network (ARLNet) for skin lesion classification. 
Zhou~et~al.~\cite{zhou2023self} employed a self-supervised pre-training approach using masked autoencoders (MAE) for medical image analysis tasks, which led to improved performance in chest X-ray disease classification tasks. The integration of additional information has been shown to considerably boost diagnostic accuracy. For instance, Wang~et~al.~\cite{wang2019kgznet} incorporated external medical knowledge to guide their training process. Similarly, Han~et~al.~\cite{han2022radiomics} proposed a radiomics-guided transformer (RGT) that combined global image information with local radiomics-guided auxiliary information to enable accurate cardiopulmonary pathology localization and classification. Despite significant advancements in various medical imaging analysis tasks~\cite{wang2019kgznet,zhang2019attention,han2022radiomics}, UTS diagnosis from KUB images is still under study. The most relevant work to date, to our knowledge, is by Liu et al.~\cite{liu2022deep} for the classification of kidney stones. However, the aforementioned methods exhibit certain limitations. Firstly, CNN-based approaches may not effectively incorporate localization information. Secondly, although attempts have been made to include location data, the subtle differences among stones, critical for accurate diagnosis, necessitate further investigation. These gaps highlight the need to develop methods that can capture fine-grained distinctions.

\begin{figure*}
  \centering
  \includegraphics[width=1\textwidth]{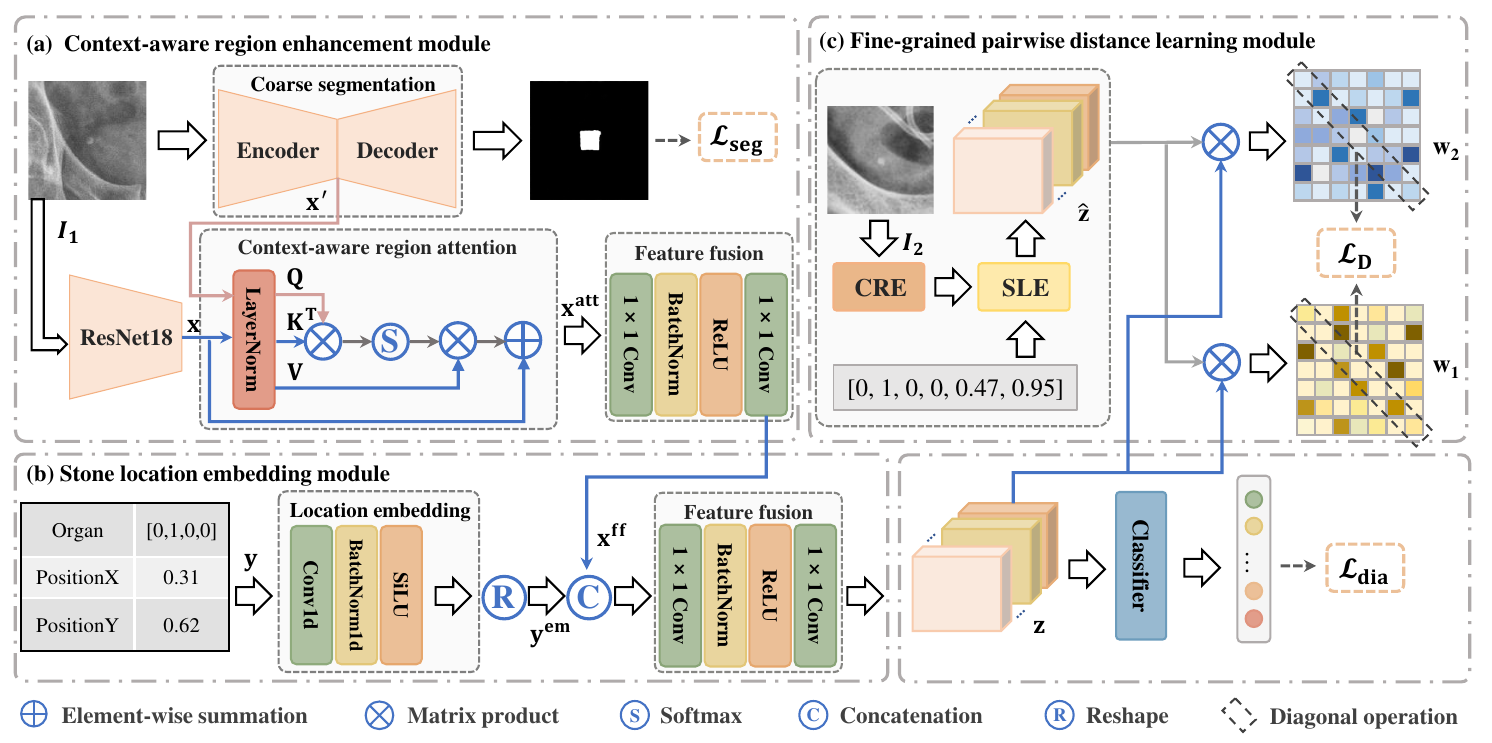}
  \caption{The overall architecture of LEPD-Net. (a) Context-aware region enhancement module. (b) Stone location embedding module. (c) Fine-grained pairwise distance learning module. It is noted that the fine-grained pairwise distance learning module will be removed during inference.}
  \label{model}
\end{figure*}

In this work, we propose the location embedding based pairwise distance learning network (LEPD-Net)~\footnote{https://github.com/BioMedIA-repo/LEPD-Net.git} for the fine-grained diagnosis of urinary stones. The detailed architecture of LEPD-Net, as depicted in Fig.~\ref{model}, consists of three pivotal components: the context-aware region enhancement (CRE) module, the stone location embedding (SLE) module, and the fine-grained pairwise distance learning (FPD) module. The CRE module identifies stone-related regions, mitigating the potential interference from high-density objects and thus enhancing the model's capability for image feature representation. Besides, the SLE module encodes textual location information, capturing crucial stone location knowledge. Finally, the FPD facilitates interactions between image pairs, enabling the recognition of fine-grained objects and improving the model's ability to discriminate between fine-grained stones in scenarios of data scarcity.

The contributions of this study are threefold. Firstly, we propose a novel LEPD-Net that integrates location information into the image modality and strengthens the interactions between image pairs, thereby enhancing the discriminative power with limited data. Second, to overcome the lack of KUB datasets for stone diagnosis, we establish an in-house dataset of 414 patients from Shuang-Ho hospital. Third, our method demonstrates consistent performance improvements over recent state-of-the-art approaches applied to both medical and natural images.

\section{Dataset and preprocessing}
\begin{figure}
  \centering
  \includegraphics[scale=0.32]{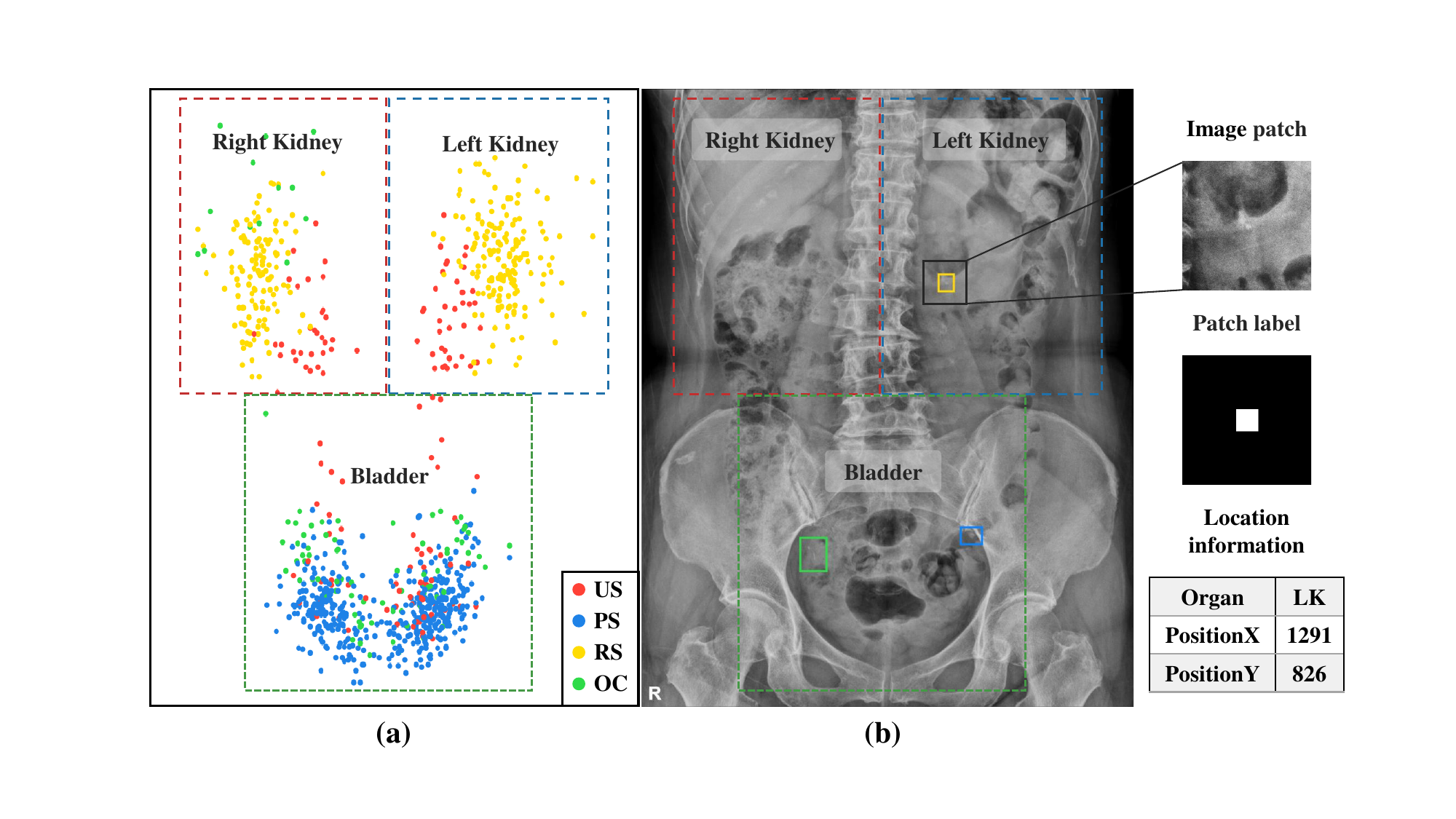}
  \caption{(a) Visualization of location distributions of 974 stones with four different types. (b) An example of typical stone patch and the location information.} 
  \label{data_introduction}
\end{figure}

\textbf{Dataset:} This retrospective study was conducted at Shuang-Ho hospital, where anonymized KUB images from 414 patients treated for UTS were collected. The data collection was approved by the institutional review and ethical board. The resolutions of these images range from 864$\times$924 to 3,311$\times$3,969 pixels. The dataset contains 974 cases of stones, including 139 cases of ureter stones (US), 448 cases of phleboliths (PS), 296 renal stones (RS), 91 other types of calcifications (OC), and 410 randomly extracted non-stone (NS) patches. All the stone patches were labeled by urologists, with each bounding box centered on the stone, as illustrated in Fig.~\ref{data_introduction}(b).

\textbf{Preprocessing:} We extract patches and location information from KUB images through a two-step process: (1) Stone patches are extracted based on the annotated bounding boxes. (2) The location information includes the coordinates and the organ region of a stone. The coordinates (Position X, Position Y) are the center of a stone's bounding box. The organ region, i.e., right kidney (RK), left kidney (LK), bladder (BL), and other regions (OR), is identified by using a stone location map generated by all stone data points, as shown in Fig.~\ref{data_introduction}. Following the collection of this information, the location data is then systematically encoded for further analysis.

\section{Methodology}
As illustrated in Fig.~\ref{model}, the LEPD-Net consists of context-aware region enhancement (CRE), stone location embedding (SLE), and fine-grained pairwise distance learning (FPD). In the CRE module, global features are first extracted using a ResNet18~\cite{he2016deep} backbone, facilitating an initial representation of the stone. Concurrently, a segmentation network utilizes the coarsely annotated bounding box around the stone as the ground truth to extract stone-related regions within the latent space. These regions are then combined with the global features through the CRE module, enhancing the model's ability to capture comprehensive representations while emphasizing critical stone-specific details. In the SLE module, location information is carefully embedded and subsequently concatenated with the features enhanced by the CRE module. Finally, the FPD module employs pairwise distance learning to identify minor variations across different stone types, thereby enhancing the model's ability for fine-grained discrimination.

\subsection{Context-aware region enhancement (CRE) module}
Given the challenge of distinguishing between stones and high-density objects with ambiguous characteristics, it is essential to approximate the localization of the stone region for precise downstream diagnosis. Inspired by the concept that segmentation can enhance diagnostic accuracy~\cite{jin2021cascade}, we adopt a coarse segmentation network and propose a context-aware region attention to enhance stone context regions, as shown in Fig.~\ref{model}(a).

\textbf{Context-aware region attention:} Mathematically, for each preprocessed patch $I_{\text{1}}$ with its corresponding ground truth $Y_{\text{1}}$, we concurrently process it through the coarse segmentation network and the pretrained ResNet18 backbone, yielding the stone-related features $\mathbf{x^{\prime}}$ and global features $\mathbf{x}$, respectively. It is important to note that the ground truth for coarse segmentation is derived from the annotated bounding box, as illustrated in Fig.~\ref{data_introduction}(b). Utilizing $\mathbf{x^{\prime}}$ and $\mathbf{x}$, we employ a context-aware region attention mechanism to investigate the interaction between stone-related features and global stone features. Specifically, we project $\mathbf{x^{\prime}}$ into a query vector $\mathbf{Q} = \operatorname{LayerNorm}(\mathbf{x^{\prime}})$, and $\mathbf{x}$ into both key $\mathbf{K}= \operatorname{LayerNorm}(\mathbf{x})$ and value $\mathbf{V}=\operatorname{LayerNorm}(\mathbf{x})$ vectors, where $\operatorname{LayerNorm}$ denotes the layer normalization function. The attention-augmented features $\mathbf{x}^{\text{att}}$ are computed as follows:
\begin{equation}
  \mathbf{x}^{\text{att}} = \operatorname{Att} (\mathbf{Q}, \mathbf{K}, \mathbf{V}) +  \mathbf{x}, \quad  \operatorname{Att} (\mathbf{Q}, \mathbf{K}, \mathbf{V}) = \operatorname{Softmax}(\frac{\mathbf{Q} \mathbf{K}^{\text{T}}}{\sqrt{d}})\mathbf{V}
\label{cra}
\end{equation}
where $\frac{1}{\sqrt{d}}$ acts as a scaling factor, $\mathbf{K}^{\text{T}}$ is the transpose of $\mathbf{K}$. Following this computation, we employ a set of feature fusion functions to integrate those features, generating the region-enhanced features $\mathbf{x}^{\text{ff}}\in\mathbb{R}^{c\times h\times w}$. Here, $c$, $h$, and $w$ represent the channel, height and width, respectively.

\subsection{Stone location embedding (SLE) module}
Given the variety of stones that may be present in the pelvis, diagnosing urinary tract stones from patch-based images is notably challenging, primarily due to the lack of global location information. We hypothesize that incorporating additional information, specifically the location and region of stones within the body, could significantly enhance visual prediction performance. This additional information, referred to as prior knowledge, is extracted during the preprocessing stage, as illustrated in Fig.~\ref{data_introduction}(b).

The `Organ' attribute is transformed into one-hot encoding matrices, and the numerical attributes, `Position X' and `Position Y', undergo normalization to fit within a [0,1] range. The resulting 6-dimensional vector $\mathbf{y}\in\mathbb{R}^{6}$ is subsequently embedded through a sequence comprising a 1D convolutional layer, a batch normalization layer, and a sigmoid linear unit (SiLU) layer. This embedded feature is then expanded to match the dimensions of the region-enhanced features $\mathbf{x}^{\text{ff}}\in\mathbb{R}^{c\times h\times w}$, yielding the location-embedded features $\mathbf{y}^{\text{em}}\in\mathbb{R}^{c\times h\times w}$. Afterwards, the location-embedded features $\mathbf{y}^{\text{em}}$ and the region-enhanced features $\mathbf{x}^{\text{ff}}$ are concatenated and fused using a feature fusion module, as depicted in Fig.~\ref{model}(b). The computation of the final feature $\mathbf{z}$ is as follows:
\begin{equation}
  \mathbf{z} = \operatorname{Fusion} (\operatorname{Concat}(\mathbf{x}^{\text{ff}},\mathbf{y}^{\text{em}})),
\label{sle}
\end{equation}
where $\operatorname{Concat}$ is the concatenation operation, the $\operatorname{Fusion}$ module consists of two 1$\times$1 convolutional layers, a batch normalization layer, and a rectified linear unit (ReLU) layer.

\subsection{Fine-grained pairwise distance learning (FPD) module}
The challenge posed by class imbalance can result in the network overfitting to sample-specific features, particularly when distinguishing visually similar classes such as PS, US, and OC. To mitigate this issue, we propose a fine-grained pairwise distance learning module, which aims to reduce the proximity of images belonging to the same stone type while increasing the distance between those of different types.

During the training stage, we randomly select two images to create an image pair ($I_{\text{1}}$, $I_{\text{2}}$) and a corresponding ground truth pair ($Y_{\text{1}}$, $Y_{\text{2}}$). These samples belonging to the same stone type are treated as positive sample pairs, i.e., $Y_{\text{1}}=Y_{\text{2}}$, whereas the two samples from different types are regarded as negative sample pairs, i.e., $Y_{\text{1}} \neq Y_{\text{2}}$.
To explore the interactions between image pairs, we first compute high-level features ($\mathbf{z}$, $\hat{\mathbf{z}}$) for the pair ($I_{\text{1}}$, $I_{\text{2}}$) through the CRE and SLE modules. Subsequently, we employ a cross-attention mechanism on these pairwise features, which allows us to derive a pair of attention weights ($\mathbf{w}_1$, $\mathbf{w}_2$). The attention weights are helpful in determining the relative importance of features within each image in relation to its counterpart. The process is formulated as follows:
\begin{equation}
  \mathbf{w}_1 = \operatorname{Softmax}(\frac{\hat{\mathbf{z}} \cdot \mathbf{z}}{\sqrt{d}}),
  \mathbf{w}_2 = \operatorname{Softmax}(\frac{\mathbf{z} \cdot \hat{\mathbf{z}}}{\sqrt{d}}),
  \label{cam}
\end{equation}
Finally, we minimize the distance $\text{D}$ between the two samples. The distance between the high-level features of the image pair is defined as follows:
\begin{equation}
  \text{D} =
      \begin{cases}
      \operatorname{Distance}(\operatorname{diag}(\mathbf{w}_1), \operatorname{diag}(\mathbf{w}_2)), & \text{if } Y_{1} = Y_{2} \\
      \max \left(0, 1 - \operatorname{Distance}(\operatorname{diag}(\mathbf{w}_1), \operatorname{diag}(\mathbf{w}_2))\right), & \text{otherwise},
      \end{cases}
\end{equation}
where the cosine similarity serves as the distance measure, denoted by $\operatorname{Distance}$, and $\operatorname{diag}$ returns a square diagonal matrix of weights. Optimizing this attention pair increases the difficulty of network training and reduces the overfitting to sample-specific features. Note that the FPD is only used for training and will be removed for inference without consuming extra computational cost.

\subsection{Loss function}
For the loss ($\mathcal{L}_{\text{dia}}$) of diagnosis, we integrate the label smoothing strategy~\cite{szegedy2016rethinking} with the distribution of stones, aiming to reduce the potential label noise introduced during the annotation process and to prevent overfitting. For coarse segmentation, the Dice coefficient is employed as the loss function ($\mathcal{L}_{\text{seg}}$). Regarding the fine-grained pairwise distance learning, the loss ($\mathcal{L}_{\text{D}}$) is calculated by directly minimizing the cosine distance $\text{D}$. Consequently, the total loss ($\mathcal{L}$) is formulated as:
\begin{equation}
  \mathcal{L} = \mathcal{L}_{\text{dia}} + \alpha \mathcal{L}_{\text{seg}} + \beta \mathcal{L}_{\text{D}}
\end{equation}
where $\alpha$ and $\beta$ are weighting coefficients that balance the contributions of each loss.



\section{Experiments and results}

\subsection{Implementation details and evaluation measures}
Our method is implemented in PyTorch using an NVIDIA RTX 3090 graphic card. To optimize our model, we use the Adam optimizer with a polynomial learning rate policy where the initial learning rate $1\times10^{-4}$ is multiplied by $(1 - \frac{epoch}{total\_{epoch}})^{power}$ with $power$ as 0.9. We set the batch size to 16, with a total training duration of 200 epochs. The hyper-parameters $\alpha$ and $\beta$ are empirically set to 0.1. Training patches are resized to 224$\times$224 after applying different online augmentations, including ColorJitter, RandomGrayscale, GaussianBlur, and RandomHorizontalFlip, to improve data variety.
It is noted that the encoder and decoder architecture employed for coarse segmentation can be substituted with other models, such as U-Net~\cite{ronneberger2015u}.
To ensure the robustness of our findings, we conduct extensive 5-fold cross-validation across all experiments.

\begin{table}
  \small
  \centering
  \caption{Urinary stones diagnosis performance of recently proposed methods.}
  \label{tab1}
  \renewcommand\arraystretch{0.89}
  \setlength{\tabcolsep}{2.4mm}
  \begin{tabular}{c|c|c|c|c}
  \hline
  Method          & Acc (\%) & Pre (\%) & F1 (\%) & Sen (\%)\\
  \hline
  ResNet18~\cite{he2016deep}              & 92.81$\pm$0.52  & 76.51$\pm$6.83  & 67.79$\pm$5.37  & 65.20$\pm$5.58  \\
  ARLNet~\cite{zhang2019attention}        & 93.48$\pm$0.35  & 76.17$\pm$2.40  & 71.38$\pm$1.59  & 68.98$\pm$2.03  \\
  MobileNetV3~\cite{howard2019searching}  & 93.28$\pm$0.34  & 78.34$\pm$2.46  & 70.97$\pm$1.54  & 67.73$\pm$2.22  \\
  SwinTransformer~\cite{liu2021swin}      & 91.60$\pm$0.28  & 65.66$\pm$6.52  & 57.08$\pm$6.01  & 56.88$\pm$6.55  \\
  Conformer~\cite{peng2021conformer}      & 92.58$\pm$0.83  & 72.95$\pm$3.68  & 67.43$\pm$5.19  & 64.98$\pm$5.47  \\
  MAE~\cite{he2022masked}                 & 88.66$\pm$1.01  & 60.42$\pm$11.33 & 49.52$\pm$7.45  & 50.44$\pm$6.35  \\
  RepLKNet-B~\cite{ding2022scaling}       & 90.88$\pm$0.64  & 65.89$\pm$1.67  & 59.84$\pm$1.20  & 57.68$\pm$1.65  \\
  SMPConv-T~\cite{kim2023smpconv}         & 93.21$\pm$0.53  & 74.67$\pm$3.12  & 69.85$\pm$3.29  & 68.16$\pm$3.62  \\
  \hline
  LEPD-Net        & \textbf{94.98$\pm$0.46} & \textbf{82.42$\pm$1.92} & \textbf{78.58$\pm$2.49} & \textbf{76.90$\pm$3.42}  \\
  \hline
  \end{tabular}
\end{table}

We employ accuracy (Acc), precision (Pre), F1 score (F1), and sensitivity (Sen) as the evaluation metrics.

\subsection{Performance comparison}
To demonstrate the effectiveness of our LEPD-Net, we implement several state-of-the-art image classification methods, which include models based on traditional CNN architectures (ResNet18~\cite{he2016deep}, ARLNet~\cite{zhang2019attention}, and MobileNetV3~\cite{howard2019searching}), transformer-based approaches (SwinTransformer~\cite{liu2021swin}, Conformer~\cite{peng2021conformer}, and MAE~\cite{he2022masked}), and large kernel-based models (RepLKNet-B~\cite{ding2022scaling} and SMPConv-T~\cite{kim2023smpconv}). To ensure a fair comparison, all models are trained under identical settings.

\begin{figure}
  \centering
  \includegraphics[scale=0.37]{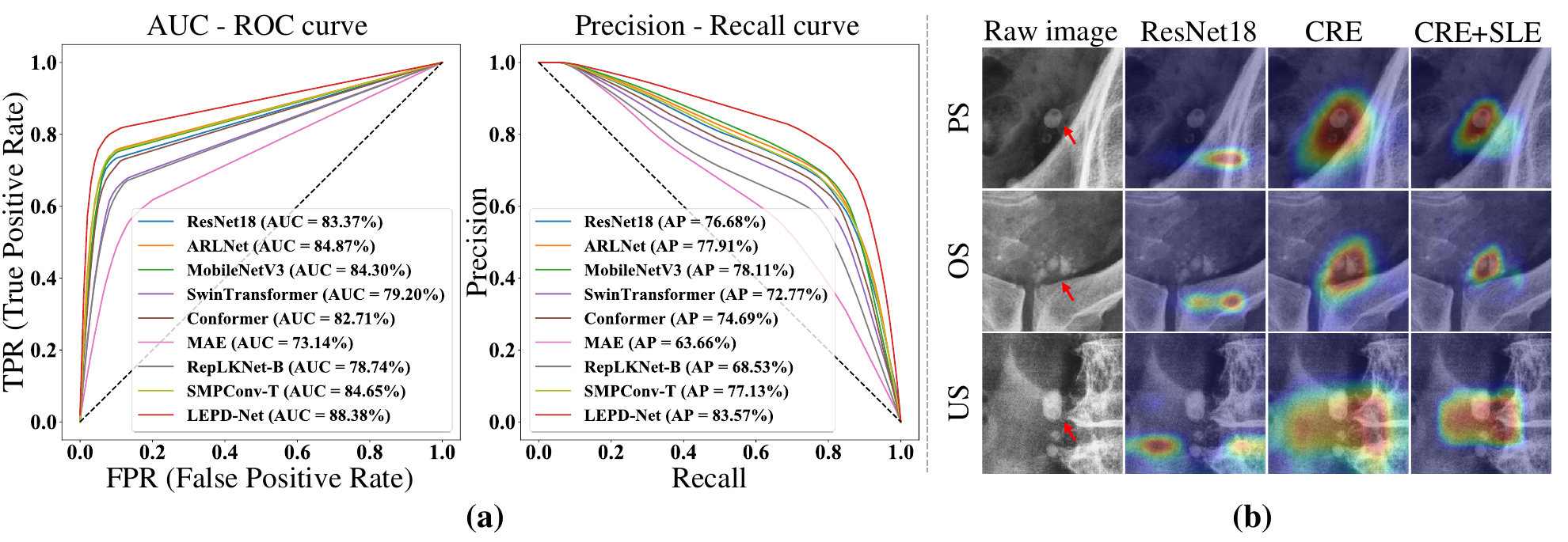}
  \caption{(a) AUC-ROC and Precision-Recall curves for our LEPF-Net and other comparing methods. (b) Visual saliency maps for challenging-to-classify stones.}
  \label{curve_result}
  \end{figure}

The results presented in Table~\ref{tab1} show the superior performance of LEPD-Net. Notably, LEPD-Net achieves an outstanding Pre of 82.42\%, which represents a 4.08\% improvement over the second-best model, MobileNetV3. Furthermore, LEPD-Net achieves an F1 of 78.58\%, surpassing MobileNetV3 by 7.61\%, demonstrating the method's balanced precision and sensitivity in classification tasks. Additionally, LEPD-Net demonstrates a significant enhancement in Sen with a score of 76.90\%. These improvements highlight the enhanced ability of LEPD-Net to accurately diagnose urinary stones.


\begin{table}
  \small
  \centering
  \caption{Urinary stones diagnosis performance with ablation studies.}
  \label{tab2}
  \setlength{\tabcolsep}{2.4mm}
  \renewcommand\arraystretch{0.89}
  \begin{tabular}{c|c|c|c|c|c|c}
  \hline
  CRE & SLE & FPD & Acc (\%)         & Pre (\%)         & F1 (\%)          & Sen (\%)         \\
  \hline
          &         &         & 92.81$\pm$0.52  & 76.51$\pm$6.83  & 67.79$\pm$5.37  & 65.20$\pm$5.58  \\
  $\surd$ &         &         & 93.70$\pm$0.63  & 81.46$\pm$2.71  & 71.69$\pm$4.83  & 68.92$\pm$4.33  \\
          & $\surd$ &         & 93.93$\pm$0.73  & 79.40$\pm$6.37  & 72.80$\pm$3.29  & 70.78$\pm$4.47  \\
          &         & $\surd$ & 93.68$\pm$0.55  & 78.88$\pm$1.59  & 71.06$\pm$1.93  & 67.11$\pm$2.32  \\
          & $\surd$ & $\surd$ & 94.13$\pm$0.47  & 79.64$\pm$3.59  & 71.46$\pm$1.71  & 68.23$\pm$1.80  \\
  $\surd$ & $\surd$ &         & 94.36$\pm$0.50  & 81.08$\pm$3.51  & 74.08$\pm$3.02  & 71.96$\pm$4.91  \\
  $\surd$ &         & $\surd$ & 94.18$\pm$0.46  & 80.55$\pm$1.82  & 72.99$\pm$3.13  & 70.48$\pm$3.43  \\
  $\surd$ & $\surd$ & $\surd$ & \textbf{94.98$\pm$0.46} & \textbf{82.42$\pm$1.92} & \textbf{78.58$\pm$2.49} & \textbf{76.90$\pm$3.42} \\

  \hline
  \end{tabular}
\end{table}

We further visualize the AUC-ROC and Precision-Recall curves to provide an intuitive demonstration of the enhanced performance. As depicted in Fig.~\ref{curve_result}(a), the proposed LEPD-Net achieves the highest AUC and the highest average precision (AP) scores, thereby validating the effectiveness of our proposed method.

Two observations can be drawn from the results. First, CNN-based methods outperform other approaches. This could be attributed to the relatively small size of the dataset, which may have a performance ceiling that CNN-based methods can more easily reach, while the more heavily parameterized transformer-based methods may be prone to overfitting. Second, the integration of CRE, SLE, and FPD modules in our LEPD-Net leads to superior performance metrics, particularly in the sensitivity score, which is crucial for identifying true positive cases.

\subsection{Ablation study}
We further perform ablation studies to evaluate the individual contributions of our newly proposed components. As shown in Table~\ref{tab2}, the inclusion of the CRE and SLE modules results in a notable enhancement, increasing the average Acc by 1.55\% over the baseline model. The addition of FPD elevates the average Sen score from 71.96\% to 76.90\%. This improvement is critical in a clinical setting, as it indicates the model's ability to correctly identify positive cases, thereby reducing the likelihood of false negatives.

We further employed class activation maps (CAMs) to visualize the class-specific discriminative regions, thereby validating the enhanced diagnostic capabilities of CRE and SLE components. As shown in Fig.~\ref{curve_result}(b), the integration of CRE and SLE enables the network to incorporate global information, significantly improving its precision in diagnosis.

\section{Conclusions}
In conclusion, we propose a location embedding based pairwise distance learning model for the fine-grained diagnosis of urinary stones. The proposed model consists of a context-aware region enhancement module, a stone location embedding module, and a fine-grained pairwise distance learning module for improving the feature representation ability of the network. Additionally, we construct an in-house annotated dataset for stone diagnosis. Comprehensive experiments demonstrate the superiority of the proposed method. Our future work includes the extension of our approach to other medical image diagnosis tasks.

\subsubsection{Acknowledgements} This work was supported by the National Natural Science Foundation of China [Grant No. 62201460, No. 62222311, and No. 62322112], the Basic Research Programs of Taicang [Grant No. TC2023JC22], and the Fundamental Research Funds for the Central Universities.

%
%
%
%
\bibliographystyle{splncs04}
\bibliography{samplepaper}




\end{document}